\pdfoutput=1

\documentclass[11pt]{article}

\usepackage{acl} % preprint, final
\usepackage{times}
\usepackage[T1]{fontenc}
\usepackage[utf8]{inputenc}
\usepackage{microtype}
\usepackage{inconsolata}
\usepackage{graphicx}
\usepackage{adjustbox}

\usepackage{amsmath}
\usepackage{amssymb}
\usepackage{booktabs}
\usepackage{dcolumn}
\usepackage{multirow}
\usepackage{hyperref}

\usepackage{xspace}

\usepackage{enumitem} \setlist{nosep, left=\parindent} 
\usepackage[ruled, vlined, noend]{algorithm2e}

\newcommand{\nrk}[1]{\textcolor{blue}{#1}}

\newcommand{\sparta}{\texttt{SpaRTA}\xspace}
\newcommand{\lora}{\texttt{LoRA}\xspace}
\newcommand{\dora}{\texttt{DoRA}\xspace}

\newcommand{\gemma}{\texttt{gemma-2b}\xspace}
\newcommand{\gemmait}{\texttt{gemma-2b-it}\xspace}
\newcommand{\mistral}{\texttt{mistral-7b}\xspace}
\newcommand{\mistralit}{\texttt{mistral-7b-it}\xspace}

\newcommand{\eat}[1]{}

\DeclareMathOperator*{\softmax}{softmax}

\title{Sparsity May Be All You Need:\\Sparse Random Parameter Adaptation}

\author{Jesus Rios, \, Pierre Dognin, \, Ronny Luss, \,  Karthikeyan Natesan Ramamurthy \\ 
IBM Research \\ 
 \{jriosal, pdognin, rluss, knatesa\}\!@us.ibm.com}

\begin{document}

\maketitle

\begin{abstract}
Full fine-tuning of large language models for alignment and task adaptation has become prohibitively expensive as models have grown in size. Parameter-Efficient Fine-Tuning (PEFT) methods aim at significantly reducing the computational and memory resources needed for fine-tuning these models by only training on a small number of parameters instead of all model parameters. Currently, the most popular PEFT method is the Low-Rank Adaptation (LoRA), which freezes the parameters of the model and introduces a small set of trainable parameters in the form of low-rank matrices. We propose simply reducing the number of trainable parameters by randomly selecting a small proportion of the model parameters to train on, while fixing all other parameters, without any additional prior assumptions such as low-rank structures. In this paper, we compare the efficiency and performance of our proposed approach to other PEFT methods as well as full parameter fine-tuning. We find our method to be competitive with LoRA when using a similar number of trainable parameters. Our findings suggest that what truly matters for a PEFT technique to perform well is not necessarily the specific adapter structure, but rather the number of trainable parameters being used.

\end{abstract}

\section{Introduction}

It has become common practice to train application-ready language models in two phases~\citep{radford2018gpt1, kenton2019bert}: first, the model is \emph{pre-trained} on a very large and general corpus of (unlabeled) text; then further trained (or \emph{fine-tuned}) on a smaller specific set of examples demonstrating the intended behavior for a particular application, such as instruction following~\citep{ouyang2022training}. 

Overall, supervised fine-tuning (SFT) requires less computational resources than pre-training (PT) due to the significantly smaller size of the training set as well as the typical use of \emph{early stopping} to deal with overfitting. This means orders-of-magnitude less gradient computations and parameter updates are needed during SFT compared to PT. 
However, a major drawback is that memory requirements remain the same, unless a parameter-efficient fine-tuning (PEFT) technique is used. The main memory bottleneck during training is the number of trainable parameters, since additional memory must be allocated for their gradients and other per-parameter statistics needed by the optimizer.
The idea behind PEFT~\citep{lialin2023scaling} is to significantly reduce the number of trainable parameters during fine-tuning while maintaining performance.

\begin{figure}[t!]
\centering
        \includegraphics[width=0.9\linewidth]{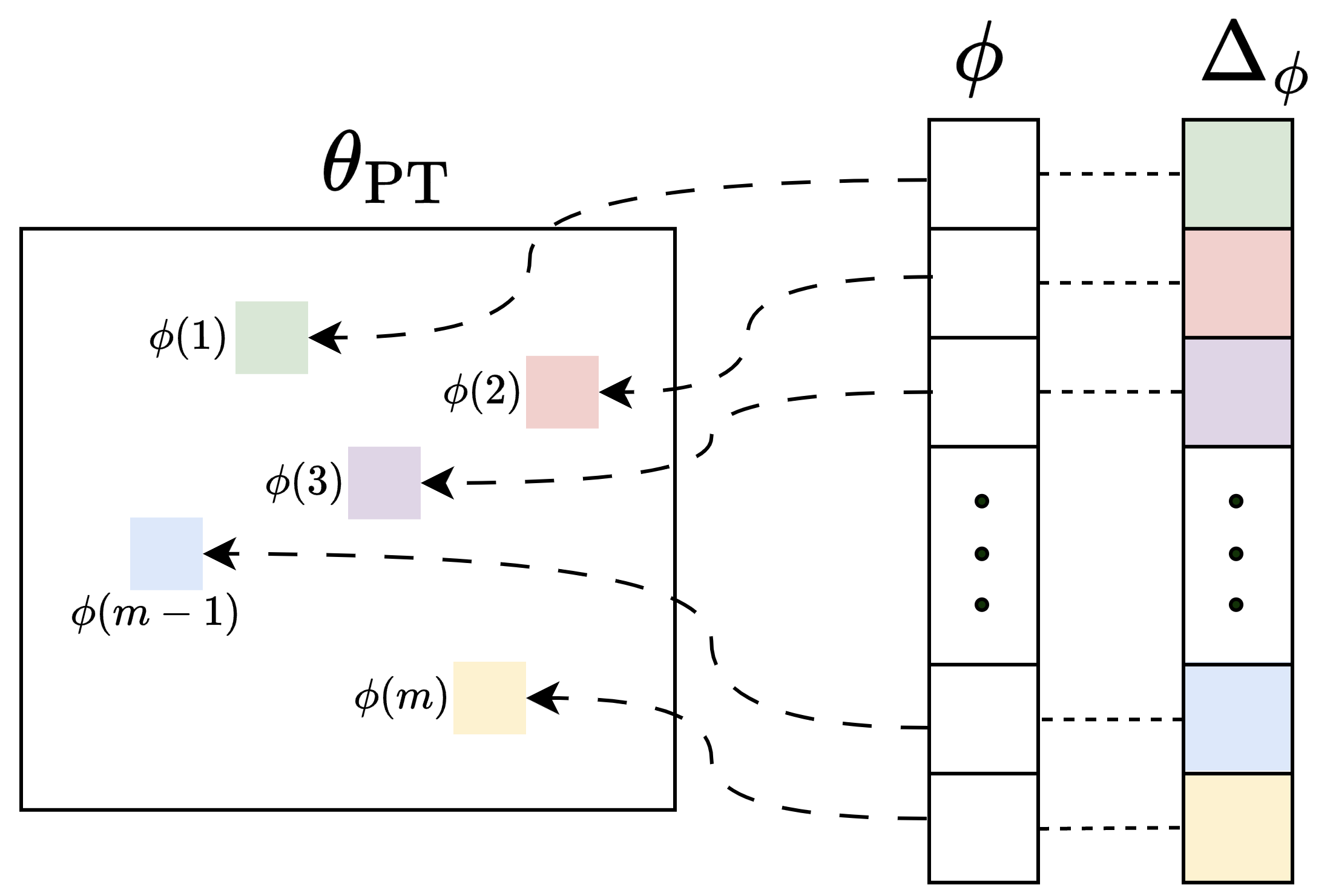}
    \caption{The proposed \sparta method which randomly chooses a small subset of parameters from $\theta_{\text{PT}}$, stores the indices of the selected parameters in $\phi$ and updates the model via adapter $\Delta_{\phi}$.}
    \label{fig:sparta}
\end{figure}

Low Rank Adaptation (\lora), first introduced by~\citet{hu2021lora}, currently remains the most popular PEFT technique. \lora freezes all the pre-trained model parameters $\theta_{\text{PT}}$ and introduces trainable low-rank matrices (e.g., $B$, $A$) % for a pre-selected subset of matrices 
to represent the changes ($\Delta = BA$) needed for adapting the model to a new task. The adapted model parameters are given by $\theta_{\text{PT}} + \Delta$. Memory and computational efficiency are achieved by optimizing only over the parameters of these newly added, but significantly smaller, matrices. 

The success of \lora begs one to ask what properties make this method perform well. Is the low-rank structure critical, i.e., does $\Delta$ need to be low-rank? Is it sufficient to constrain $\Delta$ to be low dimensional? A main goal of this paper is to investigate these questions. An abundance of research is going into new methods for structured $\Delta$ (see Section~\ref{s:related_work} below) and novel directions into unstructured methods for fine-tuning could open avenues in areas such as model merging \cite{model-soups, model-merging-fisher} or pluralistic alignment \cite{feng-etal-2024-modular}.

In this work, we propose a different approach where $\Delta$ is not factorized into low rank matrices but rather chosen to be a random subset of the model parameters. This \underline{Spa}rse \underline{R}andom parame\underline{T}er \underline{A}daptation ({\sparta}) method imposes a sparsity constraint on the adapter that can be easily controlled. By changing the desired sparsity, one can change the number of adaptation parameters. Regardless of how the selected parameters are sampled from the model parameters, subsequent updates only affect these parameters. This sparsity constraint and randomness of selected parameters is in contrast to techniques such as \lora that effectively affect all parameters in $\theta_{\text{PT}}$. See Figure~\ref{fig:sparta} for an illustration of the method. Generally, one samples $m$ parameters from the pre-trained model $\theta_\text{PT}$, stores their indices in $\phi$, and uses adapter $\Delta_\phi$ to fine-tune $\theta_\text{PT}$. 

To investigate the performance of \sparta, we build adapters with different sparsity levels and evaluate them on a wide range of natural language understanding benchmarks. \sparta is compared to other PEFT approaches including \lora and \dora \cite{dora2024}, and found to be quite competitive compared to these methods given that it only modifies a small sparse number of model parameters.

\section{Motivation}

\citet{yu2024SuperMario} look at the differences between a language model's parameters before and after fine-tuning, and demonstrate empirically that it is possible to randomly drop (i.e., set to zero) up to 99\% of these parameter changes, represented by $\Delta$, without significantly affecting model performance. 

This motivates our approach, \sparta, which produces a performant fine-tuned model by directly adapting only a small percentage of the pre-trained model parameters. \sparta randomly selects the (scalar) parameters to train and freezes the remaining parameters (i.e., setting the corresponding $\Delta$ values to zero). This $\Delta$-sparsity  also helps in reducing overfitting, as pre-trained models typically have more than enough capacity to learn the often limited amount of (labeled) data used for fine-tuning. \eat{Indeed, the redundancy in the $\Delta$ parameters observed after full fine-tuning is plausibly the result of the methods applied to prevent overfitting (\nrk{What redundancy and we need some reference}).}
There is no guarantee of $\Delta$-sparsity in \lora, but this is a desired property since it reduces parameter interference~\citep{ties-merging} when merging fine-tuned models. 

\sparta, like \lora, reduces the number of gradient computations during training (when compared with full fine-tuning), and ultimately has the same inference cost as the original model (after merging back the sparse $\Delta$ into the model). For an adaptation technique, having lower memory and computation needs during training as well as identical inference time to the original model are all quite desirable properties. \sparta has all these properties plus the unique added benefit of producing only sparse changes in a small number of the parameters of the original model that can be beneficial for merging multiple \sparta adapters.

\subsection{Is Low-Rank Adaption Necessary?} % for $\Delta$
\label{lora_justification_analysis}

In Appendix \ref{sec:need_for_lora}, we show that the changes (i.e., $\Delta$) in weight matrix during full parameter fine-tuning are, in fact, not generally low-rank for capable models such as \gemmait and \mistralit. This indicates that \lora works, not particularly because of its low-rank constraint, but rather due to the reduction in model capacity achieved by \lora  when fine-tuning on limited training data, as it is typically done in task adaption. Such insight also hints that any constraint reducing the capacity of the original model could perform competitively, motivating our approach which selects a small number of parameters from the original model to be updated during training.

\section{Related Work}
\label{s:related_work}
The last few years has seen many advances in PEFT methods. Perhaps the most well-known and used method in practice is \lora \cite{hu2021lora}, which has spurred many variants including: \dora, which adapts only the directions of pre-trained weights~\cite{dora2024}; VeRA, which shares low-rank matrices across layers~\cite{vera2024}; AdaLoRA, which adaptively allocates parameter budgets among weight matrices according to their importance~\cite{adalora2023}; and SoRA, which dynamically adjusts the intrinsic rank during adaptation~\cite{sora2023}. Each of these methods has a different structure to the $\Delta$ being optimized, with the commonality being the training of some function of low-rank matrices. AdaLoRA and SoRA are also dynamic methods, which adjust the function during adaptation.
%For example, further expanding on VeRA \cite{vera2024}, the authors propose to learn $\Delta=DBEA$ where $A$ and $B$ are random low-rank matrices and $D$ and $E$ are diagonal matrices to be trained. 

Beyond adding structured parameters is the concept of fine-tuning a small subset of the total parameters of the model, i.e., sparse fine-tuning, where one must first decide which subset of parameters to fine-tune (similar to deciding which parameters in each layer to add adapters to in \lora). \citet{ansell2024scaling} start with an initial set of parameters and offer procedures to \emph{drop} (according to parameter magnitude) and \emph{grow} (according to gradient) the set, i.e., they learn the set of parameters to train. Alternatively, \citet{ma-etal-2024-sparsity} focus on the sparsity of neuron activations during training by pre-computing neuron importance scores and only including \emph{important} neurons in computations during training. \citet{ansell-etal-2022-composable} first fine-tune on all parameters, select the parameters that change the most, and then fine-tune again from scratch on the selected parameters. %There are also successful Delta Parameter Pruning (DPP) methods such as , which sparsifies the $\Delta$ matrix after fine-tuning by random pruning following by rescaling, and 
\citet{deng2025dare} proposes improvements via a modified pruning strategy to the random dropping and rescaling method of  \citet{yu2024SuperMario} which motivates this work.
In contrast to these more complex, sometimes dynamic approaches, our proposed \sparta method produces a performant fine-tuned model by directly adapting only a small percentage of the pre-trained model parameters chosen completely at random.

Yet another related direction is that of compression which results in sparse models; algorithms in this genre take a dense fine-tuned model with the goal of compressing it while maintaining performance. Compression (see \citet{compression_survey} for a survey) could be accomplished by quantization, low-rank approximation, pruning (i.e., removing neurons, attention heads, or even layers), or distillation. The focus of this paper, however, is on fine-tuning dense models rather than learning sparse models as in \citet{sparsity-may-cry}.

\section{\sparta: Adapting a Random Subset of Model Parameters}

Suppose the parameters of a pre-trained language model are $\theta_{\text{PT}} \in  \mathbb{R}^n$ where $n$ is the number of parameters, and full parameter fine-tuning (FT) is performed with a labeled dataset characterizing a task. FT typically updates $\theta_{\text{PT}}$ using a stochastic first-order gradient-based optimization algorithm, e.g., Adam from \citet{diederik2015adam}, to maximize the conditional probability of the labels in the training dataset under the model. The FT model is then given by $\theta_{\text{FT}} = \theta_{\text{PT}} + \Delta_{\text{FT}}$ where $\Delta_{\text{FT}} \in  \mathbb{R}^n$. 

This has two drawbacks in terms of memory efficiency. First, the optimizer state can be large, e.g., the state of the Adam optimizer is $4$ times as large as the parameter space as it includes current parameter values, their gradients as well as per parameter statistics of those gradients. 
Second, storing a new FT model takes as much memory as the PT model which could be an issue when many task-specific models are requested.

\sparta proposes to \textit{randomly} select a small subset of the model parameters to optimize while freezing the rest. The model parameters $\theta \in \mathbb{R}^{n}$ are partitioned into trainable $\theta_\text{T} \in \mathbb{R}^{m}$ and frozen $\theta_\text{F} \in \mathbb{R}^{n-m}$ ones, with $m\ll n $ being the number of selected parameters. 
This allows our approach to have a drastically lower memory footprint than FT by reducing the size of the optimizer state as well as faster training by reducing the number of gradients to compute. % similar to \lora.
This is similar with respect to \lora in that both reduce the memory footprint by optimizing a small set of parameters (the adapter) while freezing most (\sparta) or all (\lora) of the model parameters.

In our \sparta implementation\footnote{Code is available at https://github.com/IBM/SpaRTA.}, we introduce: (i) non-trainable \textit{indices} $\phi \in \mathbb{R}^m$ containing the indices of the randomly selected elements in $\theta_\text{PT}$, and (ii) trainable parameters $\Delta_\phi \in \mathbb{R}^m$ representing the subset of $\Delta$ that \sparta learns at indices $\phi$. The pseudocode for \sparta is given in Algorithm~\ref{algo:sparta}. 

\begin{algorithm}[t!]
\SetAlgoLined
\KwIn{Pre-trained model with $\theta_\text{PT}\in\mathbb{R}^{n}$} 

\KwIn{Task labeled dataset $\mathcal{D}$}

\textbf{Sample $\phi$:} Indices of (scalar) parameters to be optimized 

\textbf{Initialize:} $\Delta_\phi=0$

\While{validation loss has not converged} 
{

\textbf{1. Merge:} $\theta_\phi = \theta_\phi + \Delta_\phi$ 

\textbf{2. Compute loss:} Continue forward pass from $\theta_\phi$ using a batch of labeled data from $\mathcal{D}$

\textbf{3. Compute gradients} with respect to $\Delta_\phi$ using backpropagation 

\textbf{4. Unmerge:} $\theta_\phi = \theta_\phi - \Delta_\phi$, without recording this operation in the computational graph 

\textbf{5. Update $\Delta_\phi $} with the Adam Optimizer

}
\KwOut{$\phi, \,\Delta_\phi$}
\caption{{\bf \sparta}}
  \label{algo:sparta} 
\end{algorithm}

Whereas optimizing \lora requires computing gradients with respect to the $A$ and $B$'s used to compute  $\Delta$ (which constitute additional parameters, independent from $\theta_\text{PT}$), \sparta requires computing gradients with respect to $\Delta_\phi$, the changes over a subset of parameters chosen from $\theta_\text{PT}$ and indexed by $\phi$. 
%This nuance is why \lora can trained with out-of-the-box optimizers, but \sparta requires a different implementation. For example, Step 4 in Algorithm~\ref{algo:sparta} must take into account the computational graph so as to not interfere with the forward pass to be computed at the following iteration. Specifically, this unmerge operation must not be recorded in the computational graph.

Generating the index set $\phi$ in \sparta is done by sampling from a Bernoulli independently and including each scalar parameter in $\phi$ with probability $m/n$.  
Hence, \sparta uses $m$ trainable parameters out of $n$ in expectation. 
%Step 5 also depends on the specific optimizer and settings being used; \sparta uses Adam in our implementation. 
Finally, inference is performed, similarly to \lora, by merging the fine-tuned $\Delta$ into $\theta_\text{PT}$ and then making the necessary forward passes. Thus, \sparta does not introduce any additional \emph{inference} latency. % when deploying the resulting fine-tuned model.

\section{Memory Usage} 
\label{mem_usage}

Recall that \sparta freezes $n-m$ ($m \ll n$) of the model parameters.
We define sparsity as $s = 1 - m/n \in (0, 1)$, the percentage of frozen model parameters (e.g., if $1\%$ of parameters are trainable, then the sparsity is $99\%$). Subsequently, density is defined as $k = m/n = 1 -s$, the percentage of trainable model parameters. 
In practice, for a chosen sparsity $s$, one can freeze a model parameter with probability $s$, expecting a total sparsity percentage of $s$ over all model parameters. Thus, in expectation, $k= m/n$ percent of the model parameters are chosen as trainable, for a total of $n \, k = m$ trainable parameters.

For \sparta, only $\Delta_\phi$ (of size $m$) is trainable, which is significantly smaller than the total number of model parameters since $m \ll n$. However, the indices of these randomly chosen parameters must be recorded into $\phi$, adding to the memory requirements. Indices can be stored in $16$-bit integers for all the Transformer models \citep{attention-is-all-you-need} considered in this paper.

\sparta sparsifies neither the model \emph{head} (which is kept fully trainable) nor the \emph{embeddings} (which is kept frozen during training). The parameters in Transformer networks consist of bias vectors and two-dimensional weight matrices. Storing indices for all these trainable parameters would require at most $m\,(2  \times 16)$ bits of memory ($m$ parameters, two integers to index a 2-dimensional matrix, 16 bits per integer).

The values in $\Delta_\phi$ are of the same type as the model parameters, e.g., using $16$-bit brain floating-point (\texttt{bfloat16}) tensors. Thus, \sparta requires up to $m \, (2 \times 16 + 16 )= 3 m \times 16 $ bits of extra memory to specify the index $\phi$ and delta $\Delta_\phi$ tensors. That is $ 3 \, k $ 
times more memory than the original model, which requires just $n \times 16$ bits for storing its parameters. 
For instance, using \sparta on a model with sparsity 80\%, 90\%, 95\% and 99\% would require up to 60\%, 30\%, 15\% and 3\% more memory, respectively.

% training advantage 

We next analyze\footnote{This analysis does not include memory requirements associated with model buffers (e.g., those used to track running statistics for layer normalization) because they are relatively small and the same for both the Full FT and \sparta.} 
\sparta's memory savings during training. Optimizing the full set of model parameters (FT) using Adam \citep{diederik2015adam} requires memory for the parameters, their gradients, and their (adaptive) first and second moments estimates. This requires $4\,n \times 16$ bits of memory when using a \texttt{bfloat16} representation.

In contrast, \sparta only optimizes $\Delta_\phi$, requiring a total of $ m \, (4  \times 16 +  2 \times 16) + n \times 16$ bits of memory: 
(i) $m \, (4  \times 16)$ bits needed by Adam to optimize $\Delta_\phi$ of size $m$, 
(ii) $m \, (2 \times 16)$ bits for $\phi$ with the indices identifying the model parameters associated with $\Delta_\phi$, 
and 
(iii) the PT model parameters ($n \times 16$ bits). 
Memory savings appear then if and only if
\begin{align}
   m \, (4 \!\times\! 16 +  2 \!\times\!16) + n \times 16 &< 4n \!\times\! 16 \\ 
   k n \, (6 \!\times\! 16) & < 3n \!\times\! 16 \nonumber \\
   k & < 0.5, \nonumber % \frac{1}{2}
\end{align}
that is, \sparta is a real PEFT method iff $k < 0.5$, i.e., a sparsity higher than 50\% is required ($s\!>\!0.5$), making less than 50\% of the model parameters trainable. For instance, using \sparta on a model with sparsity $s=$ 80\%, 90\%, 95\%, and 99\% requires
45\%, 60\%, 67.5\% and 73.5\%  less memory than full parameter FT, 
respectively, see Table~\ref{tab:memory_efficiency_2b}. Additional savings are possible regarding storing indices given a fixed random number generator; in this case, the random path of selected parameters to train can be derived given a fixed seed. 

\begin{table}[!ht]
\footnotesize
    \centering
    \begin{adjustbox}{max width=\columnwidth}
    \begin{tabular}{cccccccc} 
    \toprule  
      Model & storage & Full FT&  \multicolumn{5}{c}{\sparta} \\ 
      %\cmidrule(lr){1-2} 
      %\cmidrule(lr){3-3} 
      \cmidrule(lr){4-8}
     $n$ & (on disk) & 0\% & 50\% & 80\%  & 90\% & 95\% & 99\% \\
	%\midrule
    \cmidrule(lr){1-2}
    \cmidrule(lr){3-3}
    \cmidrule(lr){4-8}
    2B & 4 & 16 & 16 & 8.8 & 6.4 & 5.2  & 4.2 \\
    7B & 14 & 56 & 56 & 30.8 & 22.4 & 18.2 & 14.8  \\
    \bottomrule
    \end{tabular}
    \end{adjustbox}
    \caption{\sparta memory usage efficiency  during training for $n=\text{2B}$ and 7B parameter models. Memory in Gigabytes (GB) for storing the parameters on disk (storage) as well as for FT with Adam on (i) the full set of parameters (Full FT) equivalent to 0\% sparsity (ii) a sparse random subset (\sparta) for different sparsity percentages from 50\% to 99\%.} 
    \label{tab:memory_efficiency_2b}
\end{table}

\section{Experimental Setup}
We next detail our experimental framework. Motivation is first given for the tasks followed by a description of models to be used and how they are adapted (with manipulations) to the tasks.

\subsection{Tasks}

The main experimental focus is on Natural Language Understanding (NLU) tasks, specifically \emph{sequence classification}, which involves classifying natural language sequences into a given number of classes.
NLU tasks are easier to evaluate than Natural Language Generation (NLG) tasks, as one can use simple metrics (Accuracy, F1-scores, etc.) that can avoid the inherent ambiguity of evaluating more general generative tasks. While \sparta is also applicable to NLG tasks, they are not used in the following demonstrations due to the challenges associated with their evaluation, typically requiring human judgments as the gold standard measure of performance. Human evaluations can be expensive and time consuming, do not scale easily, and are not as \emph{objective} as NLU evaluations.

\subsection{Language Models}
\label{LMs}

Starting with available open-weight language models, the goal is to adapt them to perform a selected NLU task. %\emph{sequence classification}. 
Two types of trained models are used: \emph{base} and \emph{instruction}-tuned models, where the latter have additionally been trained to follow users' intent when prompted, such as answering questions or responding to instructions.
Specifically, we consider the following language models: \gemma and \gemmait from the Gemma family~\citep{gemma}, and \mistral and \mistralit from the Mistral\footnote{We  use v0.3 for both models.} family~\citep{mistral7b}.
%, developed by Google and Mistral AI respectively. 
The \texttt{"it"} suffix refers to instruction-tuned models. They are of particular interest for our experiments as they will show results for models with two different numbers of parameters. All models are readily available text-to-text, decoder-only transformer models with open weights, downloadable from Hugging Face. % \footnote{\url{https://huggingface.co/}}. 

Note that when using an \emph{instruction}-following (\texttt{it}) model, inputs are formatted according to the conventions established when training the model on instructions.

\subsection{Task Adaptation}
\label{task_adaption}

% input 
When using a \emph{base} model for a sequence classification task, the raw text to be classified is input directly into the model as a sequence of tokens (with some special token if necessary to deal with structured inputs). However, when using an \emph{instruction} model, these sequences are first wrapped into a classification-specific instruction. For example, a possible instruction could be: ``Determine if the following sentence has a positive sentiment. Respond Yes or No.'', followed by the sequence itself to be classified.  

% model
A (generative) pre-trained Transformer model with a decoder-only architecture has a head that transforms the final hidden state of each token in the input sequence into vocabulary logits. 
Efficiently adapting this model for sequence classification requires the swap of this head for a sequence classification head, which uses only the final hidden state of the last token in the input sequence $h \in \mathbb{R}^d$ to do the classification. 
This reduces the parameter size of the head, which is just a linear layer applied to $h$, from a weight matrix $W \in \mathbb{R}^{v \times d} $ to $W \in \mathbb{R}^{c \times d}$, where $d$ is the dimension of the model's hidden states (e.g., 2,048 and 4,096 for the \gemma and \mistral models respectively), $v$ is the number of tokens in the vocabulary (e.g., 256,000 for \gemma or 32,768 for \mistral), and $c$ is the number of classes (e.g. $2$ to $4$ in the experiments that follow). With this small change, the model outputs classification probabilities for each input sequence through 
\begin{equation} \label{model-head}
p = \softmax(h\, W^T) \in \mathbb{R}^c.
\end{equation}
 
The classification heads of our \emph{base} pre-trained models (i.e., \gemma and \mistral) are initialized with random weights. The weights of the the original vocabulary heads of instruction-tuned models (i.e., \gemmait and \mistralit)  are rather reused when initializing their classification heads. To do so requires to first identify the tokens in the vocabulary that the model is expected to use for classification following the instruction. For example, these could be the tokens associated with a ``Yes'' or ``No'' response. The embeddings in the original model (head) associated with those classification tokens are extracted and used to initialize the classification head. While many models tie their vocabulary heads to their tokens embedding matrices, these new classification heads are never tied to the model's input embedding matrix.   

\section{Experimental Results}   
\label{s:experimental_results}

Efficacy of \sparta is demonstrated empirically by comparing it to two baselines, \lora (only the additional matrices representing low-rank adapters are optimized)  and \dora (magnitude is optimized in addition to low-rank adapters).
%on both sequence classification and natural language understanding tasks. 
We also explore  the performance of \sparta on a range of sparsity levels, varying the number of trainable parameters. 

\begin{table}[t]
\centering
\begin{tabular}{lcrrr} 
 \toprule
 Dataset & Classes & Train & Dev & Test \\
 \midrule
 IMDB  &  2&   25,000 & 5,000 & 20,000 \\
% \midrule
  COLA  &  2 & 7,551 & 1,000  & 1,043  \\
  MNLI  &  3 & 100,000 & 10,000 & 19,647 \\
  MRPC  &  2 & 3,668 & 408 & 1,725  \\
  QNLI  &  2 & 99,725 & 5,000  & 5,463 \\
  QQP   &  2 & 100,000 & 5,000 & 40,430  \\
  RTE   &  2 & 2,182 & 300 & 277 \\
  SST-2 &  2&   66,349 & 1,000 &  872 \\
%  \midrule
  BoolQ &  2&   9,427 & 1,270 &   2,000\\
%\midrule
  MMLU  &  4 & 99,842 & 1,531 & 14,042 \\
\bottomrule 
\end{tabular}
\caption{Sequence classification datasets. Training sets limited to 100K samples. Training samples with > 256 tokens are removed (here using \gemma tokenizer, with \mistral tokenizer in Appendix~\ref{datasets_appendix}).}
\label{tab:datasets}
\vspace{-.1cm}
\end{table}

We consider several NLU benchmarks, including IMDB~\citep{IMDBdataset}, GLUE~\citep{glue}, BoolQ~\citep{boolq} and MMLU~\citep{mmlu}. \eat{Each sequence classification task considered in our experiments is given by a dataset of labeled examples.} See Appendix~\ref{datasets_appendix} for detailed descriptions. 
Table~\ref{tab:datasets} summarizes these datasets and our splits for training, development, and testing. A detailed description of the training setup can be found in Appendices~\ref{training_details} and~\ref{training_parameters}.  All results are averaged over 3 random seeds.

\subsection{IMDB}

\begin{table}[htbp]
  \centering
 \begin{tabular}{lcc}
    \toprule
Sparsification targets & Loss & Accuracy \\
    \midrule
%    Wq & 0.167 & 93.9\% \\
%    Wk & 0.133 & 95.6\% \\
%    \textbf{Wv} & \textbf{0.109} & \textbf{96.7\%} \\
%    \textbf{Wo} & \textbf{0.109} & \textbf{96.7\%} \\
    Wq, Wv (like LoRA) & 0.117 & 96.3\% \\
    \textbf{Wv, Wo} & \textbf{0.109} & \textbf{96.7\%} \\
    Wq, Wk, Wv & 0.123 & 95.9\% \\
    Wq, Wk, Wo & 0.112 & 96.0\% \\
    Wq, Wq, Wv, Wo & 0.110 & 96.5\% \\
    \midrule
    MLP & 0.150 & 94.7\% \\
    Wq, MLP & 0.150 & 94.5\% \\
    Wk, MLP & 0.150 & 95.1\% \\
    Wv, MLP & 0.145 & 95.2\% \\
    Wo, MLP & 0.140 & 95.6\% \\
    Wq, Wk, MLP & 0.150 & 94.6\% \\
    Wv, Wo, MLP & 0.136 & 95.8\% \\
    W, MLP, norm & 0.140 & 95.4\% \\
    \bottomrule
  \end{tabular}
    \caption{Loss and accuracy on the SST-2 task after applying our \sparta approach to different types of parameters in the \gemma model, given that the same number of trainable parameters are selected. Results are averaged across 10 random seeds.\eat{, with $\text{std}=\pm 0.02\%$ over the loss.}}
  \label{tab:ablation_study}
\end{table}

\begin{table*}[htb]
\footnotesize
\centering
\begin{adjustbox}{max width=0.85\textwidth}
\begin{tabular}{llcccccccccccc} 
 \toprule
  & & \multicolumn{2}{c}{\textbf{\gemma}} &  \multicolumn{2}{c}{\textbf{\gemmait}} &  
 \multicolumn{2}{c}{\textbf{\mistral}}  &   \multicolumn{2}{c}{\textbf{\mistralit}} \\ \cmidrule(lr){3-10}
  method & density (\%) & loss & acc. &  loss & acc. &  loss & acc. &  loss & acc. \\
 \midrule 
\textbf{PT (zero-shot)}  &              &--     &--       & 0.425 & 92.0\% & --    & --     & 0.435 & 86.1\% \\ \midrule 
\textbf{Full FT} & 100\%           & 0.092 &  96.9\% & 0.107 & 96.3\% & 0.080 & 97.4\% & 0.080 & 97.3\% \\ \midrule 
%\sparta (10\%)     & 0.094 &  96.8\% & 0.100 & 96.6\% & 0.077 & 97.4\% & 0.080 & 97.4\% \\ 
\textbf{\sparta} & 5\%      & 0.096 &  96.7\% & 0.105 & 96.3\% & 0.084 & 97.2\% & 0.081 & 97.5\% \\ 
%\sparta (1\%)      & 0.096 &  96.7\% & 0.104 & 96.4\% & 0.085 & 97.1\% & 0.083 & 97.1\% \\ 
\textbf{\sparta} & 0.5\%    & 0.096 &  96.7\% & 0.103 & 96.4\% & 0.087 & 96.9\% & 0.087 & 97.0\% \\ 
\textbf{\sparta} &  0.05\%  & 0.106 &  96.3\% & 0.114 & 96.2\% & 0.080 & 97.3\% & 0.076 & 97.4\% \\ \midrule 
\textbf{\lora}  & $\approx$0.05\%             & 0.101 &  96.5\% & 0.113 & 96.2\% & 0.086 & 97.1\% & 0.081 & 97.1\%\\ %\midrule
%Head  & $\approx$2e-4\%              & 0.165 &  93.6\% & 0.194 & 92.7\% & 0.139 & 94.9\% & 0.117 & 95.8\% \\  
\bottomrule 
\end{tabular}
\end{adjustbox}
\caption{Test loss and accuracy of models adapted to the IMDB dataset with different fine-tuning methods. We also report zero-shot performance of instruction following PT models. For training details, see Table~\ref{tab:imdb-training-params}.} 
\label{tab:imdb}
\end{table*}

%%% GEMMA 2B IT
\begin{table*}[htb]
\centering
% \footnotesize
\begin{adjustbox}{max width=\textwidth}
\begin{tabular}{lcccccccccccccccccc} 
\multicolumn{19}{c}{\textbf{\texttt{Model: \texttt{gemma-2b-it}}}} \\ 
\toprule
 & & \multicolumn{2}{c}{\textbf{QNLI}} &  \multicolumn{2}{c}{\textbf{RTE}} &  
 \multicolumn{2}{c}{\textbf{SST2}}  &   \multicolumn{2}{c}{\textbf{QQP}} &
 \multicolumn{2}{c}{\textbf{MNLI}}  &   \multicolumn{2}{c}{\textbf{MRPC}} & \multicolumn{2}{c}{\textbf{COLA}} & \multicolumn{2}{c}{\textbf{BoolQ}} &

 \\ \cmidrule{3-19}
method                     & targets & loss   & acc.  & loss   & acc.  & loss   & acc.  & loss   & acc.  & loss   & acc.  & loss   & acc.  & loss   & mcc   & loss   & acc.   & avg. \\ \midrule
\textbf{PT (zero-shot)}    &     &  1.34 &  59.5 &  1.77 &  56.3 &  0.75 &  60.7 &  1.41 &  48.3 &  2.91 &  30.9 &  0.91 &  64.6 &  0.85 &   0.7 &  0.88 &  68.5 &  44.7 \\  \midrule
\textbf{\sparta (5\%)}     & ALL  &  0.17 &  93.7 &  0.54 &  81.0 &  0.14 &  95.0 &  0.26 &  89.0 &  0.33 &  87.2 &  0.35 &  85.3 &  0.39 &  56.2 &  0.36 &  84.4 &  84.0 \\ 
\textbf{\sparta (0.5\%)}   &     &  0.16 &  94.0 &  0.43 &  80.7 &  0.15 &  94.7 &  0.24 &  89.7 &  0.32 &  87.8 &  0.34 &  86.5 &  0.45 &  54.9 &  0.37 &  84.7 &  84.1 \\ 
\textbf{\sparta (0.037\%)}  &     &  0.18 &  93.1 &  0.48 &  76.9 &  0.16 &  94.5 &  0.26 &  88.8 &  0.35 &  86.4 &  0.36 &  84.5 &  0.45 &  54.8 &  0.40 &  83.6 &  82.8 \\  \midrule
\textbf{\lora (0.037\%)}   & Q,V  &  0.18 &  93.4 &  0.44 &  78.7 &  0.14 &  95.1 &  0.26 &  89.1 &  0.33 &  87.8 &  0.33 &  85.4 &  0.41 &  55.0 &  0.36 &  84.0 &  83.6 \\ 
\textbf{\dora (0.037\%)}   &     &  0.18 &  93.4 &  0.45 &  78.0 &  0.15 &  95.6 &  0.26 &  89.0 &  0.33 &  87.8 &  0.33 &  85.5 &  0.41 &  57.1 &  0.36 &  84.4 &  83.8 \\ 
\textbf{\sparta (0.037\%)}  &     &  0.19 &  92.8 &  0.46 &  77.6 &  0.16 &  94.8 &  0.26 &  88.6 &  0.36 &  86.3 &  0.36 &  84.2 &  0.45 &  54.5 &  0.39 &  83.4 &  82.8 \\  \midrule
\textbf{\sparta (0.037\%)}  & O,V  &  0.17 &  93.5 &  0.44 &  80.7 &  0.15 &  94.9 &  0.26 &  88.9 &  0.34 &  87.2 &  0.35 &  85.7 &  0.42 &  55.9 &  0.37 &  85.0 &  84.0 \\ \bottomrule
\end{tabular}
\end{adjustbox}
\caption{Test loss and accuracy of \gemmait adapted to GLUE and BoolQ datasets with different fine-tuning methods. Standard errors can be found in Table \ref{tab:sterror-gemma-it}, see Appendix~\ref{additional_experiment_results}. For training details see Appendix~\ref{ss:training_details_glue_boolq}.} \label{tab:gemma-it}
\end{table*}

Table~\ref{tab:imdb} shows the results for IMDB, where each model is asked to classify a review as positive or negative. Each model is fine-tuned using the following adaptation methods: (i) Full parameter fine-tuning (Full FT) where all model parameters are optimized; (ii) \sparta for different density levels 5\%, 0.5\%, 0.05\%, with the last allowing \sparta to have approximately the same number of trainable parameters as \lora; (iii) \lora with rank $r=8$, equivalent to about $0.05\%$ of trainable parameters compared to the model full parameter size\eat{; (iv) Head adaptation where only the classification head is updated ($\approx$2e-4\% density)}. 

In Table~\ref{tab:imdb}, results for adaptation methods (rows) are sorted in order of descending density as to show the impact of decreasing the number of trainable parameters on the overall performance. For \gemma and \gemmait, Full FT adaptation gives (practically always) the best test loss and accuracy numbers. This is expected since all model parameters are fine-tuned. \sparta results at 5\% density are close to Full FT, even matching them for \gemmait. As the density decreases by orders of magnitude, the results slowly degrade. For 0.05\% density, \texttt{gemma} models' performances match or slightly lag performance from \lora. \eat{Head adaptation provides the worse results as one would expect since it has so few parameters to work with.} 

For \texttt{mistral} models, the trend is similar with Full FT showing best (or close to best) loss and accuracy numbers. \sparta performs well, matching and even improving upon Full FT with a density of 0.05\%. \sparta even improves over \lora for both \mistral and \mistralit at this low density. \eat{Once again, Head adaptation gives the worst results. }Overall these results are encouraging; \sparta is competitive, provides similar performance to \lora for low densities, and does not degrade the performance of full parameter fine-tuning.

\subsection{Which Parameters Should \sparta Target?}
\label{ss:ablation_study}

So far, \sparta has chosen trainable parameters from any layer of the transformer with equal probability. In contrast, \lora concentrates adaption on the key and value self-attention weight matrices.  

Now, we investigate whether it is better to concentrate the selection of sparse, trainable parameters on a few model layers or equally across all. Specifically, given a budget for the number of parameters that can be trained, which type of parameters should we target for sparsification (while freezing the remaining) to achieve best task performance when adapting the sparsified parameters within the targeted types?

We conduct an ablation study to answer this question. The \gemma model is adapted to the SST-2 task using our \sparta approach targeting different combinations of parameter types. We set a budget of 1.25M trainable parameters within the \gemma model so that sparsity is $s=99.95\%$. 

Results of these experiments are shown in Table~\ref{tab:ablation_study}. Concentrating the selection of trainable parameters through sparsification in the self-attention value (Wv) or/and output (Wo) weight matrices yields the best performance under the given budget.  In contrast, distributing our sparsification across different combinations of weight types may lead to significantly lower performance. 

Corresponding results confirming, as done in \citet{hu2021lora}, that targeting Wo and Wv is suboptimal for \lora can be found in Appendix \ref{additional_experiment_results}. Note that concentrating our sparsification over targeted parameter tensors, although proven beneficial in terms of single-task adaption performance, can decrease performance when merging since it increases the chance of parameter interference.

\subsection{GLUE and BoolQ}
 
%%% MISTRAL 7B IT
\begin{table*}[htb]
\centering
\centering
% \footnotesize
\begin{adjustbox}{max width=\textwidth}
\begin{tabular}{lcccccccccccccccccc} 
\multicolumn{19}{c}{\textbf{\texttt{Model: %\texttt{Mistral-7B-Instruct-v0.3}}}} \\ 
\texttt{mistral-7b-it}}}} \\ 
\toprule
 & & \multicolumn{2}{c}{\textbf{QNLI}} &  \multicolumn{2}{c}{\textbf{RTE}} &  
 \multicolumn{2}{c}{\textbf{SST2}}  &   \multicolumn{2}{c}{\textbf{QQP}} &
 \multicolumn{2}{c}{\textbf{MNLI}}  &   \multicolumn{2}{c}{\textbf{MRPC}} & \multicolumn{2}{c}{\textbf{COLA}} & \multicolumn{2}{c}{\textbf{BoolQ}} & 

 \\ \cmidrule{3-19}
method                     & targets & loss   & acc.  & loss   & acc.  & loss   & acc.  & loss   & acc.  & loss   & acc.  & loss   & acc.  & loss   & mcc   & loss   & acc.   & avg. \\ \midrule
\textbf{PT (zero-shot)}    &     &  1.38 &  76.6 &  1.54 &  66.4 &  0.66 &  66.3 &  0.85 &  57.5 &  2.42 &  35.5 &  0.70 &  70.0 &  0.72 &  25.2 &  0.39 &  85.2 &  49.6 \\  \midrule
\textbf{\sparta (5\%)}     & ALL  &  0.11 &  96.0 &  0.26 &  89.8 &  0.12 &  95.9 &  0.22 &  90.7 &  0.23 &  91.6 &  0.32 &  87.3 &  0.36 &  62.6 &  0.24 &  90.8 &  88.1 \\ 
\textbf{\sparta (0.5\%)}   &     &  0.11 &  95.8 &  0.30 &  90.5 &  0.12 &  96.0 &  0.23 &  90.4 &  0.24 &  91.2 &  0.32 &  88.0 &  0.35 &  67.9 &  0.24 &  90.5 &  88.8 \\ 
\textbf{\sparta (0.048\%)}  &     &  0.12 &  95.7 &  0.31 &  89.2 &  0.12 &  96.1 &  0.23 &  90.5 &  0.24 &  91.1 &  0.31 &  88.2 &  0.34 &  67.3 &  0.26 &  89.5 &  88.4 \\  \midrule
\textbf{\lora (0.048\%)}   & Q,V  &  0.12 &  95.6 &  0.24 &  91.2 &  0.12 &  96.0 &  0.23 &  90.2 &  0.24 &  91.0 &  0.29 &  89.3 &  0.32 &  69.9 &  0.24 &  90.5 &  89.2 \\ 
\textbf{\dora (0.048\%)}   &     &  0.13 &  95.0 &  0.27 &  90.9 &  0.11 &  96.4 &  0.23 &  90.2 &  0.25 &  90.9 &  0.28 &  88.7 &  0.32 &  68.2 &  0.24 &  90.5 &  88.8 \\ 
\textbf{\sparta (0.048\%)}  &     &  0.12 &  95.6 &  0.31 &  88.3 &  0.13 &  95.6 &  0.23 &  90.1 &  0.25 &  90.9 &  0.30 &  88.6 &  0.35 &  65.2 &  0.26 &  89.7 &  88.0 \\  \midrule
\textbf{\sparta (0.048\%)}  & O,V  &  0.11 &  95.9 &  0.31 &  88.4 &  0.14 &  95.8 &  0.23 &  90.4 &  0.24 &  91.2 &  0.30 &  88.4 &  0.34 &  64.9 &  0.26 &  89.7 &  88.1 \\ \bottomrule
\end{tabular}
\end{adjustbox}
\caption{Test loss and accuracy of the \mistralit model adapted to GLUE and BoolQ datasets with different fine-tuning methods. Standard errors can be found in Table \ref{tab:sterror-mistral-it}, see Appendix~\ref{additional_experiment_results}. For training details see Appendix~\ref{ss:training_details_glue_boolq}.} \label{tab:mistral-it}
\end{table*}

\begin{table*}[htb]
\footnotesize
\centering
\begin{adjustbox}{max width=0.85\textwidth}
\begin{tabular}{lccccc} 
 \toprule
 \textbf{method} & \textbf{targets} & \textbf{\gemmait} &  \textbf{\mistralit} & \textbf{\gemma} & \textbf{\mistral} \\ \midrule
\textbf{\lora} & Q, V & 83.6& \textbf{89.2}& 81.7& \textbf{87.3}\\
\textbf{\sparta} & Q, V & 82.8 & 88.0 &78.3 &86.7 \\
\textbf{\sparta} & O, V & \textbf{84.0}& 88.1& \textbf{83.2}& 85.6\\
\bottomrule
\end{tabular}
\end{adjustbox}
\caption{Average accuracy (\%) per adapted model. Averages are taken across 8 tasks (7 GLUE and BoolQ) based on Tables \ref{tab:gemma-it}, \ref{tab:mistral-it}, \ref{tab:gemma-pt}, and \ref{tab:mistral-pt}. \sparta is shown to be competitive with \lora, particularly when targeting Wo (O) and Wv (V) self-attention matrices, which were previously shown to be optimal for \sparta.} 
\label{tab:acc_summary}
\end{table*}

\begin{table*}[htb]
\footnotesize
\centering
\begin{adjustbox}{max width=0.85\textwidth}
\begin{tabular}{lccccc} 
 \toprule
 \textbf{method} & \textbf{targets} & \textbf{\gemmait} &  \textbf{\mistralit} & \textbf{\gemma} & \textbf{\mistral} \\ \midrule
\textbf{\lora} & Q, V & 37.58& \textbf{62.5}& 37.5& \textbf{50}\\
\textbf{\sparta} & Q, V & 0 & 0 & 0 & 31.3 \\
\textbf{\sparta} & O, V & \textbf{62.5}& 37.5& \textbf{62.5}& 18.8\\
\bottomrule
\end{tabular}
\end{adjustbox}
\caption{Win rates (\%) per adapted model,  computed across 8 tasks (7 GLUE tasks and BoolQ) based on Tables \ref{tab:gemma-it}, \ref{tab:mistral-it}, \ref{tab:gemma-pt}, and \ref{tab:mistral-pt}. \sparta is shown to be competitive with \lora, particularly when targeting Wo (O) and Wv (V) self-attention matrices, which were previously shown to be optimal for \sparta.} 
\label{tab:win_rates}
\end{table*}

\begin{table*}[htb]
\footnotesize
\centering
\begin{adjustbox}{max width=\textwidth}
\begin{tabular}{lcccccccc} 
 \toprule
 \multicolumn{9}{c}{\textbf{MMLU}}\\
 \midrule 
  &  \multicolumn{2}{c}{\textbf{\gemma}} &  \multicolumn{2}{c}{\textbf{\gemmait}} &  
 \multicolumn{2}{c}{\textbf{\mistral}}  &   \multicolumn{2}{c}{\textbf{\mistralit}} \\ 
               &  loss & acc. &  loss & acc. &  loss & acc. &  loss & acc. \\
 \midrule 
\textbf{PT (zero-shot)}     & -         & -             & 5.284 & 35.3\% & -       & -            & 1.839 & 59.3\% \\
\textbf{\sparta} &  1.249 &  45.1\%  &1.250& 45.1\% & 0.987 & 61.5\%   &  0.930& 63.1\% \\ 
\textbf{\lora}     &  1.233  & 46.9\%   &1.271& 45.1\% & 0.981 & 62.8\% &  0.928 & 63.0\% \\ 
\bottomrule 
\end{tabular}
\end{adjustbox}
\caption{Test loss and accuracy of models adapted to the MMLU dataset with SpaRTA and LoRA. We also report zero-shot performance of instruction following PT models. For training details see Table~\ref{tab:mmlu-training-params} in Appendix~\ref{training_parameters}.}
% We did multiple runs and observed a $std=0.02$ for the loss and $std=0.4$ for accuracy.
\label{tab:mmlu}
\end{table*}

We now focus on comparing \sparta and \lora methods using the 7 NLU tasks in the GLUE benchmark as well as BoolQ. Table~\ref{tab:gemma-it} presents the results for adapting \gemmait. An equivalent set of results for the \gemma model is given in Table~\ref{tab:gemma-pt}, see Appendix~\ref{additional_experiment_results}.
Adaptation results (rows) are grouped by which type of parameters are targeted, with \sparta further ordered in descending order of density. \sparta with 0.037\% density has approximately the same number of trainable parameters as \lora, for which a rank $r=8$ was used.

\eat{Overall, the same trend as for IMDB is observed. Full FT results are very often the best, sometimes bested by \sparta 5\% (RTE, MNLI). For a low density, where \lora and \sparta use the same number of trainable parameters, \sparta can best \lora (QNLI, SST2, RTE) but is overall competitive with \lora results.}

When targeting all parameters, \sparta overall exhibits better performance with higher density, i.e, when more parameters are chosen to be trained, with a few exceptions. Targeting the Wq and Wv self-attention matrices shows very similar performance between \lora, \dora, and \sparta. Targeting the Wo and Wv self-attention matrices within \sparta, which was shown to be the optimal in Section~\ref{ss:ablation_study}, yields better performance for \sparta when compared to both \lora and \dora on four of the eight datasets (QNLI, RTE, MRPC, BoolQ). 

A similar set of results on GLUE and BoolQ is provided for \mistralit in Table~\ref{tab:mistral-it}. Again, results for the \mistral model can be found in Table~\ref{tab:mistral-pt} in Appendix~\ref{additional_experiment_results}. \sparta exhibits comparable performance to \lora and \dora for a low density of 0.0048\%, where all methods have comparable numbers of trainable parameters. And once again, with \sparta targeting Wv and Wo typically outperforming the targeting of Wq and Wv.

To further illustrate the competitiveness of \sparta with \lora, Table \ref{tab:acc_summary} summarizes accuracy results across the 8 datasets while Table \ref{tab:win_rates} offers another perspective by considering win-rates among the competing methods for each model. Note that \lora exhibits best performance overall when targeting the Wq and Wv self-attention matrices as mentioned in the original \lora paper \cite{hu2021lora}, and hence we target these parameters when using \lora. While we have already explored optimal targets for \sparta, we include results targeting Wq and Wv for completeness. Across both tables, \sparta (Wo, Wv) outperforms \sparta (Wq, Wv) in six out of the eight scenarios. \lora and \sparta are equally performant on average across the two tables, demonstrating the claim that \sparta is competitive with \lora.

\subsection{MMLU}

We have chosen MMLU because it is known for being a very challenging NLU task. Table~\ref{tab:mmlu} compares \sparta against \lora on adapting each of our models to the MMLU multiple-choice question answering task, where each model must predict the correct answer to a set of questions. Solving this task requires a high level of world knowledge and problem solving skills. In this experiment, we restrict both methods to use approximately the same number of trainable parameters. See Table~\ref{tab:mmlu-training-params} in the Appendix for training details. Here again, the results show that \sparta is competitive with \lora, providing similar performance.

\subsection{Remarks}

Our results establish that \sparta is a viable adaptation technique that can be competitive with \lora, especially for larger LMs. These results demonstrate that a simple sparsifying scheme can offer a valid adaptation technique. This opens the possibility for further investigating sparse adaptation as the low-rank approximation of a model's parameter changes is not the only mechanism to provide a performant adaptation method. This indicates that what matters is not necessarily the adapter structure used in PEFT techniques, but rather the number of trainable parameters, that is relevant to the adaptation task at hand.

\section{Conclusion}
\label{ref:concl}
As PT language models have grown in size, PEFT has become crucial for enabling fine-tuning large PT language models on limited hardware and financial resources. We have introduced \sparta, an approach that sharply decreases the set of trainable parameters, reducing the GPU memory used by the optimizer and speeding up training. We have demonstrated on a variety of task adaptation scenarios that our fine-tuning approach is parameter-efficient and competitive with \lora, the current PEFT standard. Experiments with 2B and 7B parameter pre-trained models demonstrate good performance, and as per \citet{yu2024SuperMario}, we expect larger models to allow for higher levels of sparsity in training, meaning that efficiency of \sparta  should get better with larger model sizes as also suggested in Table~\ref{tab:memory_efficiency_2b}. 

Regarding future directions, while \sparta has been applied to \emph{supervised learning}, it is also amenable to \emph{reinforcement learning} often used for model alignment \citep{ouyang2022training}. We also plan to explore the merging of \sparta adapters due to their potential for little interference. 

%\sparta allows one to select which types of parameters to sparsify and at what layers. Like \lora, this decision affects both the number of trainable parameters and the performance of the fine-tuned model. 

\section{Limitations} 
We have demonstrated various benefits of \sparta, including low memory and high performance. Regarding limitations, questions remain about how to best deal with overfitting, though we have some insights. We have observed in our experiments that as we increase the sparsity level, and reduce, in turn, the number of trainable parameters, there is less overfitting. Indeed, there is a point in which both the training and validation losses converge together without significantly diverging from each other, eliminating the need for explicit overfitting mitigation techniques. Moreover, further increasing of the sparsity level beyond this point results in underfitting. Thus, we can think of our approach as a technique to improve generalization by limiting a model's capacity to overfit to the training data. However, finding the breaking point at which this happens requires expensive experimentation. We leave as future work the investigation of how such an optimal sparsity level depends on the model, dataset sizes, and task complexity. Knowing this relation will allow us to determine in advance how much sparsity in the trainable parameters is needed for reducing the capacity of a large model to a point where it learns a new task on a relatively small amount of examples without overfitting.

\bibliography{refs} % main.bbl

\appendix

\section{Ranks of Differences in Fine-Tuned Weight Matrices}
\label{sec:need_for_lora}

\lora is based on the idea that the changes in the model parameters after adapting it to a new task can have a low-rank approximation. To see if this is the case, we can easily check this presumption in a few well-known fine-tuned models by comparing their weight matrices before and after fine-tuning. Thus, for every weight matrix $W_{\text{PT}}$ (i.e.\ a $2$-dimensional trainable tensor) in the \emph{pre-trained} model, we compute the \emph{delta} matrix $\Delta = W_{\text{FT}} - W_{\text{PT}}$, where $W_{\text{FT}}$ is the corresponding weight matrix after \emph{fine-tuning} the model over all the original model parameters using standard supervised (or reinforcement) learning. \lora assumes that these matrices don't change or their differences given by $\Delta$ are low-rank.

We compute all delta weight matrices for two well-known \emph{instruction-following} fine-tuned models: \gemmait and \mistralit, see Section~\ref{LMs}. As we can see in Tables~\ref{tab:delta_rank_gemma-2b} and~\ref{tab:delta_rank_mistral-7b}, the feedforward (MLP)  \emph{delta} $\Delta$-matrices associated with each layer of the transformer network are all full rank\footnote{We compute the rank of a matrix as the number of singular values that are greater than zero, given a specified tolerance, using \texttt{torch.linalg.matrix\_rank}. We did some sensitivity analysis to determine such tolerance.} in both models. This is also the case for all self attention key (k) and value (v) projection matrices. However, the self attention query (q) and output (o) $\Delta$-matrices all show relatively small rank deficiencies: between $9$ and $2$ for the query projection matrices and between $46$ and $3$ for the output projection matrices, out of a potential maximum rank of $2,048$ for the \gemmait model; and between $1,788$ and $28$ for the query projection matrices and between $246$ and $10$ for the output projection matrices, out of a potential maximum rank of $4,096$ for the \mistralit model. The token embedding $\Delta$-matrices of both the \gemmait and \mistralit models are full rank. Of the two models, \mistralit is the only one that does not tie its token embeddings' weights to its head; thus the \mistralit's model head (untied lm head) is updated independently during fine-tuning, and the resulting delta change of this matrix is also full rank. Basically, we observe that the fine-tuning changes in the weight matrices of these well known models are all full rank, with the only exception being the changes in the query (q) and output (o) projection matrices that show small rank deficiencies. 
  
The observed ranks suggest that constraining the (delta) changes in weight matrices to be low-rank is not essential for fine-tuning models efficiently.

\begin{table*}[p]
    \centering
    \begin{tabular}{lcccc} 
    	\toprule         
	Weight matrix & dims & Layer(s) & Rank(s) & Rank deficiencies \\   % rank deficiency: the difference between the lesser of the number of rows and columns, and the rank.
	\midrule
	tokens embed & [ 256000,   2048 ]   &  - & 2048     & 0 (full rank) \\
	\midrule % W_q, W_k, W_v, W_o, W_{mlp}
	self attn q proj   & [ 2048,   2048 ] & 0 - 17  & 2046 -  2039 &  2 - 9\\
%	                                                     && 0  &2039   & 9 \\
%		                                            && 15& 2041& 7\\
%	                                                    && 14& 2042& 6\\	
%	                                                    && 6& 2044& 4\\	                                                    
%	                                                    && 1,3-5,7,8,10-13,16,17& 2045 & 3\\
%	                                                    && 2,9& 2046 & 2\\
	\midrule
	self attn k proj   &[ 256, 2048 ]   & 0 - 17  &256     &0 (full rank)\\
	\midrule
	self attn v proj   &[ 256, 2048 ]   & 0 - 17  &256     &0 (full rank)\\
	\midrule
	self attn o proj   &[ 2048, 2048 ]  & 0 - 17 & 2045 - 2002 & 3 - 46\\
%							 && 0  & 2002 & 46\\
%	                                                    && 16 & 2040 &  8\\	
%	                                                    && 11,12,14& 2042 &  6\\	      
%	                                                    && 6,8,10,13,15& 2043 &  5\\
%	                                                    && 3-5,7,9,17& 2044 &  4\\                                                                                                 
%	                                                    && 1,2  & 2045 &  3\\
        \midrule 
	mlp gate proj     &[ 16384,   2048 ] & 0 - 17  &2048   & 0 (full rank)\\
	\midrule
	mlp up proj        &[ 16384,   2048 ] & 0 - 17  &2048   & 0 (full rank)\\
	\midrule
	mlp down proj   & [ 2048,  16384 ] & 0 - 17  &2048   & 0 (full rank)\\
    	\bottomrule
    \end{tabular}
    \caption{Rank of the change differences in \gemmait model weight matrices after full fine-tuning.}
    \label{tab:delta_rank_gemma-2b}
\end{table*}

\begin{table*}[p]
    \centering
    \begin{tabular}{lcccc} 
    	\toprule         
	Weight matrix & dims & Layer(s) & Rank(s) & Rank deficiencies \\   % rank deficiency: the difference between the lesser of the number of rows and columns, and the rank.
	\midrule
	tokens embed & [ 32768,   4096 ]   &  - & 4096     & 0 (full rank) \\
	\midrule % W_q, W_k, W_v, W_o, W_{mlp}
	self attn q proj   & [ 4096, 4096 ]  &  0 - 31    & 4068 - 2308 & 28 - 1788 \\
	\midrule
	self attn k proj   &[ 1024, 4096 ]   & 0 - 31  &1024     &0 (full rank)\\
	\midrule
	self attn v proj   &[ 1024, 4096 ]   & 0 - 31  &1024     &0 (full rank)\\
	\midrule
	self attn o proj   &[ 4096, 4096 ]  & 0 - 31  & 4086 - 3850 & 10 - 246\\
        \midrule 
	mlp gate proj     &[ 14336,   4096 ] & 0 - 31  &4096   & 0 (full rank)\\
	\midrule
	mlp up proj        &[ 14336,   4096 ] & 0 - 31  &4096   & 0 (full rank)\\
	\midrule
	mlp down proj   & [ 4096,  14336 ] & 0 - 31  &4096   & 0 (full rank)\\	
	\midrule
	untied lm head& [ 4096, 32768 ]  & - & 4096  &  0 (full rank)\\
    	\bottomrule
    \end{tabular}
    \caption{Rank of the change differences in \mistralit model weight matrices after full fine-tuning.}
    \label{tab:delta_rank_mistral-7b}
\end{table*}
   
\section{Additional Dataset Details}   
\label{datasets_appendix}

\textbf{IMDB} contains a sample of ``highly polar'' movie reviews obtained from the online Internet Movie Database (IMDb) website.  IMDb registered users provide a rating (from $1$ to $10$) with each review. The reviews are binary (positive/negative) labeled with their sentiment, defined from user ratings~\citep{IMDBdataset}. Reviews with a rating higher or equal than $7$ are given a positive label; and a negative label if the rating is lower or equal than $4$. No reviews with ratings beyond these ranges are present in the dataset, which was constructed to have an equal number of positive and negative reviews, so guessing randomly yields 50\% accuracy. 

\textbf{GLUE} datasets are described in details in~\citet{glue}.

\textbf{BoolQ}~\cite{boolq} is a reading comprehension dataset with binary yes/no questions. Each example is a triplet of (passage, question, answer). It represents a natural language inference task in which a passage and a question are given as input, and a yes or no answer should be predicted. 

\textbf{MMLU} \cite{mmlu} is a benchmark designed to measure the knowledge a model acquires during pre-training. To facilitate the elicitation of a model's knowledge using multiple-choice questions, MMLU also comes with a dataset of examples on multiple-choice question answering. Thus, this training set is not designed to increase the model knowledge about the world, but to teach a model to answer knowledge questions in multiple-choice format in a zero-shot setting.

%Table~\ref{tab:datasets_app} shows the actual splits used in our experiments for the IMDB and GLUE datasets, where the training data was filtered using the \mistral tokenizer by a threshold in the maximum token length of an network input. The only differences are for the training splits of the MRPC, QNLI, and RTE datasets.

\section{Training Details}
\label{training_details}
   
We adapted each (generative) pre-trained model in Section~\ref{LMs} to do sequence classification as per the examples in any of the datasets from Table~\ref{tab:datasets}, using different supervised fine-tuning methods, including \sparta. Base models were adapted by switching their vocabulary heads to randomly initialized sequence classification heads with $c$ output classification tokens. For instruction models, we re-used the vocabulary heads as described in Section~\ref{task_adaption}. 
   
The examples (e.g.,\ text extracts, sentences) to be classified were converted into sequences of tokens before passing them as inputs to a model. We wrapped each example into an instruction to take advantage of instruction-following models, adding to the token length of the model inputs. We tried to keep such instructions as short as possible while still achieving a good initial performance before fine-tuning. For instance, after tokenization, the maximum token length of a training input from the SST-2 dataset was 
\begin{itemize}
\item 67 (without) and 87 (with instruction) for the \gemmait;   
\item 72 (without) and 95 (with instruction) for \mistralit.   
\end{itemize}
We observe here how the Gemma's tokenizer compresses more the input than Mistral's because of its larger vocabulary size; 256,000 (Gemma) vs. 32,767 (Mistral).

We tokenized all training examples before starting the fine-tuning and looked at the histogram of their token lengths. To avoid batches with too much padding and improve training efficiency, we dropped those training examples with a disproportionate large token length, i.e., corresponding to a tail of extreme values in the histogram. We only do this for the training data since evaluation (on the development or test data) requires much less compute and memory (no gradients need to be calculated and stored) and is performed less frequently. We indicate in Tables~\ref{tab:datasets} the final splits after filtering the training data this way, with the final number of training examples used for the fine-tuning. 

With \sparta, we froze the token embeddings layer, made fully trainable the classification head, and randomly chose a sparse proportion of (scalar) parameters to be trained in all the other layers of a model. We demonstrated our method with varying density levels. The total average number of trainable parameters in each case was approximately:
$99$M (5\% density), $10$M (0.5\%) and $800$k (0.037\%) for the (base and instruct) Gemma 2B models; and  
$349$M (5\% density), $35$M (0.5\%) and $3.5$M (0.048\%) for the (base and instruct) Mistral 7B models.

For \lora, we factorized the changes in the query and value self-attention projection weight matrices with rank $r = 8$ decomposition matrices, which were optimized while keeping all other model parameters frozen. The number of new trainable parameters introduced by the \lora approach in the Gemma 2B and Mistral 7B models were approximately $925$k and $3.5$M, respectively. Also, we set $\alpha = 16$ to scale the \lora adapters.

We observed overfitting when training with Full parameter FT: the development loss started deteriorating after a few epochs (e.g., approximately $2$ for SST-2) while the training loss went quickly to zero. We used a combination of early stopping, dropout and weight decay to deal with overfitting. We noticed that \sparta is a natural regularizer: increasing sparsity resulted in less overfitting, to a point in which there was no more overfitting (e.g., this was achieved at a sparsity $s>99\%$ for all models under consideration). In general, overfitting was more noticeable with the Mistral 7B models; as expected since they are larger than the Gemma 2B models. We also noticed that larger models require higher sparsity levels to eliminate overfitting given the same training data. % On the other hand, Head adapation always resulted in significant underfitting, making this fine-tuning method always underperform.

% We also observed that \sparta achieves a speedup over Full FT for sparsity levels greater than 90\%. Thus, at some sparsity level between 80\% and 90\%, computing gradients on a smaller amount of trainable parameters starts to offset the extra compute of adding and subtracting the deltas before and after the forward pass respectively. These extra operations require slicing the original parameter tensors over the randomly selected indices. 

\section{Additional Experiment Results}
\label{additional_experiment_results}

Table~\ref{tab:lora_targets} confirms targeting Wo and Wv is not the best choice for \lora (96.2\% accuracy for Wo and Wv versus 96.4\% for Wq and Wv which is the \lora default). These results can be compared with the corresponding \sparta results in Section \ref{ss:ablation_study}. Tables~\ref{tab:gemma-pt} and Table~\ref{tab:mistral-pt} present the complete set of results on GLUE and BoolQ for the \gemma and \mistral models, respectively. Tables \ref{tab:sterror-gemma-it} and \ref{tab:sterror-mistral-it} present standard errors from corresponding experiments. 

\begin{table}[htbp]
  \centering
 \begin{tabular}{lccc}
    \toprule
\lora targets & rank & Loss & Accuracy \\
    \midrule
%    Wq & 0.167 & 93.9\% \\
%    Wk & 0.133 & 95.6\% \\
%    \textbf{Wv} & \textbf{0.109} & \textbf{96.7\%} \\
%    \textbf{Wo} & \textbf{0.109} & \textbf{96.7\%} \\
Wq &16 &0.155 & 94.8 \%\\
Wk &24&0.182 & 94.5\%\\
Wv&24&0.116 & 96.2\%\\
Wo&16 &0.114 & 96.2\%\\
Wq, Wk&8 &0.152 & 94.8\%\\
Wq, Wv (default)&8 &0.110 & 96.4\%\\
Wq, Wo&6 &0.107 & 96.6\%\\
Wk, Wv&12 &0.110 & 96.7\%\\
Wk, Wo&8 &0.110 & 96.9\%\\
Wv, Wo&8 &0.114 & 96.2\%\\
Wq, Wv, Wo&4 &0.108 & 96.7\%\\
Wk, Wv, Wo&6 &0.113 & 96.7\%\\
Wq, Wk, Wv, Wo&3 &0.108 & 96.8\%\\
MLP&1 & 0.112& 96.2\%\\
\bottomrule
  \end{tabular}
    \caption{Test loss and accuracy on SST-2 after applying \lora to different types of parameters in the \gemma model, given that the same number of trainable parameters are selected. Results averaged across 10 random seeds.\eat{, with $\text{std}=\pm 0.02\%$ over the loss.}}
  \label{tab:lora_targets}
\end{table}

%%%% GEMMA-2B 
\begin{table*}[htb]
\centering
% \footnotesize
\begin{adjustbox}{max width=\textwidth}
\begin{tabular}{lccccccccccccccccc} 
\multicolumn{18}{c}{\textbf{\texttt{Model: \texttt{gemma-2b}}}} \\ 
\toprule
 & & \multicolumn{2}{c}{\textbf{QNLI}} &  \multicolumn{2}{c}{\textbf{RTE}} &  
 \multicolumn{2}{c}{\textbf{SST2}}  &   \multicolumn{2}{c}{\textbf{QQP}} &
 \multicolumn{2}{c}{\textbf{MNLI}}  &   \multicolumn{2}{c}{\textbf{MRPC}} & \multicolumn{2}{c}{\textbf{COLA}} & \multicolumn{2}{c}{\textbf{BoolQ}} 
 \\ \cmidrule{3-18}
Method               & Targets      & loss   & acc.  & loss   & acc.  & loss   & acc.  & loss   & acc.  & loss   & acc.  & loss   & acc.  & loss   & mcc   & loss   & acc.   \\ \midrule
%\textbf{PT (zero-shot)}    & ALL   &  0.70 &  50.2 &  0.70 &  51.5 &  0.70 &  46.6 &  0.68 &  55.2 &  1.11 &  33.7 &  0.71 &  45.0 &  0.71 &   0.0 &  0.71 &  48.0 \\
\textbf{\sparta (5\%)}     &  ALL   &  0.15 &  94.5 &  0.50 &  80.5 &  0.12 &  96.6 &  0.23 &  90.2 &  0.39 &  85.5 &  0.38 &  84.6 &  0.36 &  62.6 &  0.32 &  86.7 \\
\textbf{\sparta (0.5\%)}   &       &  0.15 &  94.5 &  0.51 &  77.3 &  0.12 &  96.1 &  0.23 &  90.1 &  0.42 &  84.1 &  0.37 &  84.3 &  0.36 &  62.3 &  0.34 &  85.4 \\
\textbf{\sparta (0.037\%)} &       &  0.20 &  91.9 &  0.62 &  66.3 &  0.13 &  95.4 &  0.27 &  88.3 &  1.34 &  49.3 &  0.51 &  76.1 &  0.49 &  50.7 &  0.39 &  82.8 \\ \midrule
%\textbf{\lora (0.037\%)}    &      &  0.15 &  94.2 &  0.62 &  65.2 &  0.11 &  96.8 &  0.26 &  88.9 &  0.42 &  84.0 &  0.47 &  77.3 &  0.40 &  60.9 &  0.34 &  85.7 \\ \midrule

\textbf{\sparta (0.037\%)} & Q,V   & 0.18 &  93.0 &  0.58 &  70.4 &  0.13 &  95.7 &  0.26 &  88.8 &  1.04 &  56.8 &  0.43 &  80.3 &  0.41 &  56.9 &  0.36 &  84.5 \\
\textbf{\lora (0.037\%)}    &       & 0.15 &  94.2 &  0.62 &  65.5 &  0.11 &  96.8 &  0.26 &  88.9 &  0.42 &  84.0 &  0.47 &  77.3 &  0.40 &  60.9 &  0.34 &  85.7 \\
\textbf{\dora (0.037\%)}    &       & 0.15 &  94.4 &  0.62 &  66.5 &  0.11 &  96.6 &  0.25 &  89.0 &  0.41 &  84.4 &  0.47 &  77.6 &  0.39 &  60.6 &  0.43 &  78.9 \\ \midrule

\textbf{\sparta (0.037\%)} & O,V   &  0.16 &  93.8 &  0.55 &  73.2 &  0.11 &  96.5 &  0.25 &  89.3 &  0.51 &  80.4 &  0.39 &  84.3 &  0.39 &  61.9 &  0.34 &  86.1 \\
\bottomrule
\end{tabular}
\end{adjustbox}
\caption{Test loss and accuracy of model \gemma adapted to GLUE and BoolQ datasets with different fine-tuning methods. Results are averaged over 3 random seeds. For training details see Appendix~\ref{ss:training_details_glue_boolq}.} \label{tab:gemma-pt}
\end{table*}

%%% MISTRAL 7B 
\begin{table*}[htb]
\centering
\centering
% \footnotesize
\begin{adjustbox}{max width=\textwidth}
\begin{tabular}{lccccccccccccccccc} 
\multicolumn{18}{c}{\textbf{\texttt{Model: %\texttt{Mistral-7B-v0.3}}}} \\ 
\texttt{mistral-7b}}}} \\ 
\toprule
 & & \multicolumn{2}{c}{\textbf{QNLI}} &  \multicolumn{2}{c}{\textbf{RTE}} &  
 \multicolumn{2}{c}{\textbf{SST2}}  &   \multicolumn{2}{c}{\textbf{QQP}} &
 \multicolumn{2}{c}{\textbf{MNLI}}  &   \multicolumn{2}{c}{\textbf{MRPC}} & \multicolumn{2}{c}{\textbf{COLA}} & \multicolumn{2}{c}{\textbf{BoolQ}} 

 \\ \cmidrule{3-18}
Method               & Targets      & loss   & acc.  & loss   & acc.  & loss   & acc.  & loss   & acc.  & loss   & acc.  & loss   & acc.  & loss   & mcc   & loss   & acc.   \\\midrule
% \textbf{PT (zero-shot)}    & ALL   &  0.72 &  49.6 &  0.73 &  47.6 &  0.69 &  52.3 &  0.74 &  47.3 &  1.14 &  33.0 &  0.68 &  56.4 &  0.70 &   0.0 &  0.70 &  52.5 \\
\textbf{\sparta (5\%)}     &  ALL    &  0.11 &  95.9 &  0.36 &  85.0 &  0.11 &  96.8 &  0.38 &  82.1 &  0.30 &  89.2 &  0.33 &  86.7 &  0.35 &  64.6 &  0.25 &  90.1 \\
\textbf{\sparta (0.5\%)}   &       &  0.12 &  95.8 &  0.44 &  84.4 &  0.11 &  96.7 &  0.22 &  90.6 &  0.26 &  90.4 &  0.32 &  87.0 &  0.35 &  66.8 &  0.26 &  89.9 \\
\textbf{\sparta (0.048\%)} &       &  0.13 &  95.3 &  0.41 &  83.6 &  0.11 &  96.7 &  0.23 &  90.1 &  0.40 &  86.7 &  0.34 &  86.6 &  0.36 &  64.6 &  0.26 &  89.8 \\ \midrule
% \textbf{LoRA (0.048\%)}    &       &  0.12 &  95.4 &  0.39 &  85.4 &  0.12 &  96.5 &  0.23 &  90.1 &  0.28 &  89.7 &  0.32 &  87.0 &  0.38 &  64.5 &  0.27 &  89.6 \\ \midrule

\textbf{\sparta (0.048\%)} & Q,V   &  0.14 &  94.5 &  0.37 &  84.0 &  0.11 &  96.6 &  0.24 &  89.7 &  0.33 &  88.6 &  0.36 &  85.5 &  0.36 &  65.2 &  0.28 &  89.3 \\
\textbf{\lora (0.048\%)}    &       &  0.12 &  95.4 &  0.39 &  85.4 &  0.12 &  96.5 &  0.23 &  90.1 &  0.28 &  89.7 &  0.32 &  87.0 &  0.38 &  64.5 &  0.27 &  89.6 \\
\textbf{\dora (0.048\%)}    &       &  0.12 &  95.6 &  0.41 &  85.3 &  0.12 &  96.8 &  0.24 &  90.1 &  0.28 &  89.7 &  0.33 &  86.4 &  0.34 &  65.8 &  0.26 &  89.9 \\ \midrule

\textbf{\sparta (0.048\%)} & O,V   &  0.15 &  94.5 &  0.41 &  85.6 &  0.21 &  91.7 &  0.24 &  89.6 &  0.33 &  87.8 &  0.35 &  86.4 &  0.39 &  59.8 &  0.29 &  89.3 \\

\bottomrule
\end{tabular}
\end{adjustbox}
\caption{Test loss and accuracy of model \mistral adapted to GLUE and BoolQ datasets with different fine-tuning methods.  Results are averaged over 3 random seeds. For training details see Appendix~\ref{ss:training_details_glue_boolq}.} \label{tab:mistral-pt}
\end{table*}

%%% GEMMA 2B IT STD ERROR
\begin{table*}[htb]
\centering
% \footnotesize
\begin{adjustbox}{max width=\textwidth}
\begin{tabular}{lccccccccccccccccc} 
\multicolumn{18}{c}{\textbf{\texttt{Model: \texttt{gemma-2b-it}}}} \\ 
\toprule
 & & \multicolumn{2}{c}{\textbf{QNLI}} &  \multicolumn{2}{c}{\textbf{RTE}} &  
 \multicolumn{2}{c}{\textbf{SST2}}  &   \multicolumn{2}{c}{\textbf{QQP}} &
 \multicolumn{2}{c}{\textbf{MNLI}}  &   \multicolumn{2}{c}{\textbf{MRPC}} & \multicolumn{2}{c}{\textbf{COLA}} & \multicolumn{2}{c}{\textbf{BoolQ}} 

 \\ \cmidrule{3-18}
Method                     & targets & loss   & acc.  & loss   & acc.  & loss   & acc.  & loss   & acc.  & loss   & acc.  & loss   & acc.  & loss   & mcc   & loss   & acc.   \\ \midrule
% \textbf{PT (zero-shot)}    &     &  0.00 & 0.00 &  0.00 & 0.00 &  0.00 & 0.00 &  0.00 & 0.00 &  0.00 & 0.00 &  0.00 & 0.00 &  0.00 & 0.00 &  0.00 & 0.00 \\  \midrule
\textbf{\sparta (5\%)}     & ALL  &  0.00 & 0.14 &  0.03 & 1.54 &  0.01 & 0.51 &  0.00 & 0.21 &  0.01 & 0.26 &  0.01 & 0.48 &  0.00 & 0.88 &  0.01 & 0.54 \\ 
\textbf{\sparta (0.5\%)}   &     &  0.00 & 0.16 &  0.03 & 0.64 &  0.01 & 0.46 &  0.00 & 0.17 &  0.00 & 0.13 &  0.01 & 0.07 &  0.01 & 0.75 &  0.01 & 0.25 \\ 
\textbf{\sparta (0.037\%)}  &     &  0.00 & 0.12 &  0.01 & 0.55 &  0.00 & 0.17 &  0.00 & 0.18 &  0.00 & 0.14 &  0.00 & 0.45 &  0.01 & 0.44 &  0.00 & 0.09 \\  \midrule
\textbf{\lora (0.037\%)}   & Q,V  &  0.00 & 0.11 &  0.01 & 1.10 &  0.00 & 0.35 &  0.00 & 0.03 &  0.00 & 0.11 &  0.00 & 0.22 &  0.02 & 1.40 &  0.01 & 0.12 \\ 
\textbf{\dora (0.037\%)}   &     &  0.00 & 0.11 &  0.01 & 0.55 &  0.00 & 0.20 &  0.00 & 0.03 &  0.00 & 0.09 &  0.00 & 0.32 &  0.01 & 1.31 &  0.01 & 0.29 \\ 
\textbf{\sparta (0.037\%)}  &     &  0.00 & 0.08 &  0.00 & 1.04 &  0.01 & 0.11 &  0.00 & 0.00 &  0.00 & 0.26 &  0.00 & 0.22 &  0.02 & 2.83 &  0.00 & 0.41 \\  \midrule
\textbf{\sparta (0.037\%)}  & O,V  &  0.00 & 0.13 &  0.01 & 0.64 &  0.00 & 0.27 &  0.00 & 0.16 &  0.00 & 0.17 &  0.02 & 0.22 &  0.01 & 4.36 &  0.00 & 0.22 \\ 
\bottomrule
\end{tabular}
\end{adjustbox}
\caption{Standard errors of test loss and accuracy of model \gemmait adapted to GLUE and BoolQ datasets with different fine-tuning methods.} \label{tab:sterror-gemma-it}
\end{table*}

%%% MISTRAL 7B IT STD ERROR
\begin{table*}[htb]
\centering
% \footnotesize
\begin{adjustbox}{max width=\textwidth}
\begin{tabular}{lccccccccccccccccc} 
\multicolumn{18}{c}{\textbf{\texttt{Model: %\texttt{Mistral-7B-Instruct-v0.3}}}} \\ 
\texttt{mistral-7b-it}}}} \\ 
\toprule
 & & \multicolumn{2}{c}{\textbf{QNLI}} &  \multicolumn{2}{c}{\textbf{RTE}} &  
 \multicolumn{2}{c}{\textbf{SST2}}  &   \multicolumn{2}{c}{\textbf{QQP}} &
 \multicolumn{2}{c}{\textbf{MNLI}}  &   \multicolumn{2}{c}{\textbf{MRPC}} & \multicolumn{2}{c}{\textbf{COLA}} & \multicolumn{2}{c}{\textbf{BoolQ}} 

 \\ \cmidrule{3-18}
Method                     & targets & loss   & acc.  & loss   & acc.  & loss   & acc.  & loss   & acc.  & loss   & acc.  & loss   & acc.  & loss   & mcc   & loss   & acc.   \\ \midrule
%\textbf{PT (zero-shot)}    &     &  0.00 & 0.00 &  0.00 & 0.00 &  0.00 & 0.00 &  0.00 & 0.00 &  0.00 & 0.00 &  0.00 & 0.00 &  0.00 & 0.00 &  0.00 & 0.00 \\  \midrule
\textbf{\sparta (5\%)}     & ALL  &  0.00 & 0.09 &  0.01 & 0.24 &  0.01 & 0.26 &  0.00 & 0.03 &  0.00 & 0.08 &  0.01 & 0.47 &  0.01 & 1.91 &  0.00 & 0.16 \\ 
\textbf{\sparta (0.5\%)}   &     &  0.00 & 0.05 &  0.01 & 1.07 &  0.01 & 0.18 &  0.00 & 0.10 &  0.00 & 0.09 &  0.02 & 0.25 &  0.01 & 1.28 &  0.00 & 0.09 \\ 
\textbf{\sparta (0.048\%)}  &     &  0.00 & 0.10 &  0.03 & 0.55 &  0.00 & 0.20 &  0.00 & 0.03 &  0.00 & 0.11 &  0.01 & 0.32 &  0.02 & 1.87 &  0.00 & 0.20 \\  \midrule
\textbf{\lora (0.048\%)}   & Q,V  &  0.00 & 0.16 &  0.01 & 0.60 &  0.01 & 0.44 &  0.00 & 0.06 &  0.00 & 0.13 &  0.01 & 0.33 &  0.01 & 1.16 &  0.00 & 0.17 \\ 
\textbf{\dora (0.048\%)}   &     &  0.02 & 0.76 &  0.02 & 0.87 &  0.00 & 0.17 &  0.00 & 0.03 &  0.00 & 0.07 &  0.00 & 0.17 &  0.01 & 1.22 &  0.00 & 0.16 \\ 
\textbf{\sparta (0.048\%)}  &     &  0.00 & 0.06 &  0.01 & 0.73 &  0.00 & 0.23 &  0.00 & 0.04 &  0.00 & 0.05 &  0.01 & 0.43 &  0.01 & 1.17 &  0.00 & 0.27 \\  \midrule
\textbf{\sparta (0.048\%)}  & O,V  &  0.00 & 0.02 &  0.01 & 0.42 &  0.01 & 0.34 &  0.00 & 0.07 &  0.00 & 0.13 &  0.02 & 0.24 &  0.01 & 0.97 &  0.00 & 0.23 \\ 
\bottomrule
\end{tabular}
\end{adjustbox}
\caption{Standard errors of test loss and accuracy of model \mistralit adapted to BLUE and BoolQ datasets with different fine-tuning methods.} \label{tab:sterror-mistral-it}
\end{table*}

\section{Training Hyper-Parameters}
\label{training_parameters}

The hyper-parameters used for investigating \sparta and other adaptation methods are summarized next. 

\subsection{IMDB}

The sets of best parameters used in training the various adaptation methods for each model are given in Table~\ref{tab:imdb-training-params}.
To improve training efficiency, we excluded training examples exceeding 384 tokens in length, resulting in the use of only 19,306 examples for \gemma, 18,744 examples for \gemmait, 18,373 examples for \mistral, and 17,672 examples for \mistralit for training. 
That is 77\%, 75\%, 73\%, and 71\% of the original training data of 25,000 examples, respectively.

%%% GEMMA 2B IT
\begin{table*}[htb]
\centering
\centering
% \footnotesize
\begin{adjustbox}{max width=\textwidth}
\begin{tabular}{lcccccccc} 
\multicolumn{9}{c}{\textbf{\texttt{Model: \texttt{gemma-2b-it}}}} \\ 
\toprule
 Method & \textbf{QNLI} &\textbf{RTE} &  
 \textbf{SST2}  & \textbf{QQP} &
 \textbf{MNLI} & \textbf{MRPC} & \textbf{COLA} & \textbf{BoolQ} 

 \\ \cmidrule{2-9}
\textbf{\sparta (5\%)}  & 1e-5 & 5e-4 & 5e-5 & 5e-5 & 1e-5 & 2e-4 & 5e-5 & 5e-5 \\
\textbf{\sparta (0.5\%)}  & 2e-4 & 1e-3 & 2e-4 & 2e-4 & 2e-4 & 5e-4 & 1e-3 & 2e-4 \\
\textbf{\sparta (0.037\%)}  & 1e-3 & 1e-3 & 1e-3 & 1e-3 & 1e-3 & 1e-3 & 1e-3 & 1e-3 \\
\textbf{\lora (0.037\%)} & 2e-4 & 1e-3 & 5e-4 & 5e-4 & 5e-4 & 2e-4 & 5e-4 & 5e-4 \\
\textbf{\dora (0.037\%)} & 2e-4 & 1e-3 & 5e-4 & 5e-4 & 5e-4 & 2e-4 & 5e-4 & 5e-4 \\
\textbf{\sparta (0.037\%)}  & 5e-4 & 1e-3 & 5e-4 & 5e-4 & 5e-4 & 5e-4 & 1e-3 & 5e-4 \\
\textbf{\sparta (0.037\%)}  & 5e-4 & 1e-3 & 5e-4 & 5e-4 & 5e-4 & 5e-4 & 1e-3 & 5e-4 \\
\bottomrule
\end{tabular}
\end{adjustbox}
\caption{Optimal learning rates for \gemmait model for different fine-tuning methods.} \label{tab:lr-gemma-it}
\end{table*}

%%% MISTRAL 7B IT
\begin{table*}[htb]
\centering
\centering
% \footnotesize
\begin{adjustbox}{max width=\textwidth}
\begin{tabular}{lcccccccc} 
\multicolumn{9}{c}{\textbf{\texttt{Model: % \texttt{Mistral-7B-Instruct-v0.3}}}} \\ 
\texttt{mistral-7b-it}}}} \\ 
\toprule
 Method & \textbf{QNLI} &\textbf{RTE} &  
 \textbf{SST2}  & \textbf{QQP} &
 \textbf{MNLI} & \textbf{MRPC} & \textbf{COLA} & \textbf{BoolQ} 

 \\ \cmidrule{2-9}
\textbf{\sparta (5\%)}  & 1e-5 & 5e-6 & 5e-6 & 1e-5 & 1e-5 & 5e-5 & 5e-5 & 5e-6 \\
\textbf{\sparta (0.5\%)}  & 2e-4 & 1e-4 & 5e-5 & 5e-4 & 2e-4 & 5e-4 & 2e-4 & 5e-5 \\
\textbf{\sparta (0.048\%)}  & 5e-4 & 5e-4 & 5e-4 & 1e-3 & 5e-4 & 1e-3 & 1e-3 & 1e-3 \\
\textbf{\lora (0.048\%)} & 2e-4 & 1e-4 & 2e-4 & 2e-4 & 1e-4 & 2e-4 & 2e-4 & 1e-4 \\
\textbf{\dora (0.048\%)} & 2e-4 & 1e-4 & 2e-4 & 2e-4 & 1e-4 & 2e-4 & 2e-4 & 1e-4 \\
\textbf{\sparta (0.048\%)}  & 2e-4 & 1e-4 & 2e-4 & 5e-4 & 2e-4 & 1e-3 & 1e-3 & 5e-4 \\
\textbf{\sparta (0.048\%)}  & 2e-4 & 1e-4 & 2e-4 & 5e-4 & 2e-4 & 1e-3 & 1e-3 & 5e-4 \\
\bottomrule
\end{tabular}
\end{adjustbox}
\caption{Optimal learning rates for \mistralit model for different fine-tuning methods.} \label{tab:lr-mistral}
\end{table*}

\begin{table*}[p]
\centering
\begin{adjustbox}{max width= \textwidth}
\begin{tabular}{llcccc} 
 \toprule
 & Parameter& \gemma & \gemmait & \mistral & \mistralit  \\
 \midrule
 Full FT  & batch size   & 32 & 32 & 36 &  36 \\
        & num epochs     & 2  & 2 & $\ast$ &  $\ast$ \\
        & learning rate  & 1e-5 & 1e-5 & 3e-6 & 3e-6 \\
        & max grad norm  & 10 & 50 & 120 & 120 \\
        & dropout        & 0.1 & 0.1 & 0.15 & 0.15 \\
        & weight decay   & 0.1 & 0.1 & 0.01 & 0.01 \\
 \midrule
   \sparta  & batch size  &  40   &  40  & 16 & 16 \\   
            & num epochs.  & 2     & 2    & $\ast$ &  $\ast$ \\ 
% $d=$ 10\%   & learning rate & 1e-5   & 4e-6& 2e-6  & 2e-6 \\ 
$d=$ 5\%    & learning rate & 1.5e-5 & 8e-6&  2e-6 & 2e-6 \\
%$d=$ 1\%    & learning rate & 5e-5  & 5e-5&  3e-6 & 3e-6 \\
$d=$ 0.5\%  & learning rate & 1e-4  & 5e-5& 3e-6  & 3e-6 \\
$d=$ 0.05\% & learning rate & 6e-5  & 6e-5 & 6e-5  & 6e-5 \\ 
% 	            &max grad norm & 10  &  10 & 50 & 50 \\
%	            &dropout             & 0.0 & 0.0  & 0.0  &  0.0 \\
% 	            &weight decay    & 0.0 & 0.0 & 0.0  & 0.0   \\	    
 \midrule	              
  \lora  & batch size & 40       & 40    & 20 & 20 \\
        & num epochs    &  3   &   3    & 3 & 3 \\    
  		& learning rate & 2e-4 & 2e-4 & 5e-6 & 5e-6  \\
	      & max grad norm & 15   & 15  & - &  - \\
	      & dropout       & 0.1  & 0.1    & 0.1  &  0.1\\
		& $r$           & 8    &  8   &  8  &  8 \\
		& $\alpha$      &16 & 16   & 16 & 16 \\ 
  \midrule
  Head & batch size   &  40  & 40 & 16 & 16 \\ 
       & num epochs&  4    &  4 & 3 & 3 \\ 
       & learning rate & 2e-4& 2e-4  & 1e-4 &  1e-4 \\
\bottomrule 
\end{tabular}
\end{adjustbox}
\caption{(IMDB) Training parameters used with each fine-tuning method and model in Table~\ref{tab:imdb}.  An $\ast$ in number of epochs indicates early stopping was used. For \sparta, parameters for density 5\%, 0.5\% 0,05\% are reported.} 
\label{tab:imdb-training-params}
\end{table*}

\subsection{GLUE and BoolQ}
\label{ss:training_details_glue_boolq}
For GLUE and BoolQ, training was done using a simple grid search over the learning rates: [1e-3, 5e-4, 2e-4, 1e-4, 5e-5, 1e-5, 5e-6]. The optimal learning rates found are given in Tables \ref{tab:lr-gemma-it} and \ref{tab:lr-mistral}. Weight decay was set to 0. Dropout was set to 0.0 for \sparta and 0.1 for \lora and \dora. Batch sizes were set according to what could fit in GPU memory (different per dataset and model). The number of epochs was set to 3 for instruct models and 2 for base models. Similarly to IMDB, any sample longer than 256 tokens (which is dependnent on the tokenizer used) is discarded from the dataset to avoid having few minibatch with one very long sample compared to the rest, as seen in Table~\ref{tab:datasets}.
\eat{
\begin{table*}[p]
\centering
\begin{tabular}{llc} 
 \toprule
 & Parameter & Grid Search Value Range \\
 \midrule
 Full FT & batch size & [32, 24, 8] depending on task and model to fit in GPU memory \\
    & learning rate & [1e-3, 5e-4, 2e-4, 1e-4, 5e-5, 1e-5, 5e-6]  \\
    % & max grad norm & [ None ] \\
    & dropout       &  [0.1] \\
    & weight decay  &  [0.1] \\ \midrule
 \sparta  & batch size & [same as for Full FT] \\ 
    & learning rate & [same as for Full FT] \\
    % & max grad norm & [ None ] \\
    % & dropout        & [0.0] \\
    & weight decay   & [0.1] \\	    
 \midrule	              
  \lora  & batch size & [same as for Full FT] \\ 
     & learning rate & [same as for Full FT] \\
     & $r$           &  [8] \\
     & $\alpha$      &  [16] \\
     & dropout       & [0.1] \\
     & weight decay  & [0.1] \\ 
 \midrule
 Head & batch size & [same as for Full FT] \\ 
   & learning rate & [same as for Full FT] \\
   % & max grad norm & \\
   & dropout       & [0.1] \\
   & weight decay  & [0.0]  \\
 \bottomrule 
\end{tabular}
\caption{(GLUE) Training parameters and their grid search value range to get results for each adaptation method for each model in Table~\ref{tab:gemma-it}, Table~\ref{tab:mistral-it}, Table~\ref{tab:gemma-pt}, and Table~\ref{tab:mistral-pt}. Best set of parameters were chosen from best validation loss results over 2 epochs of training.} \label{tab:glue-training-params}
\end{table*}
 }
 
\subsection{MMLU}

As with other datasets, we started approximating the deltas of the self-attention matrices Wq and Wv with rank $r=8$ matrices, resulting in approximately 1M trainable parameters for the Gemma 2B models and 3.5M for the Mistral 7B models under LoRA ($\alpha = 16$). We choose the sparsity of our SpaRTA adapter accordingly so both methods end up with the same number of parameters to train. Thus, our SpaRTA approach uses a $s=99.96\%$ for the Gemma models (i.e. making only $0.04\%$ of the original parameters  trainable); and $s= 99.95\%$ for the Mistral models.

Since we observed that \gemma struggled to learn with both \sparta and \lora methods, we decided to increase the number of trainable parameters when adapting \gemma to MMLU. Specifically, we increased the rank of the \lora adaption matrices to $r=16$, which lead to approximately 2M trainable parameters; and chose accordingly a sparsity $s=99.92\%$ for \sparta.

The MMLU dataset has a small fraction of training examples that are extremely long. We enforce a maximum input token length of $520$ for training efficiency.  This reduces the number of training examples from $99,842$ to $84,296$ and $74,100$ for the Gemma and Mistral instruction models; and to $91,321$ and $85,820$ for their respective base models. Test and validations sets are not affected by this decision. The training parameters used for adapting each model to MMLU are shown in Table~\ref{tab:mmlu-training-params}.

\begin{table*}[htbp]
\centering
\begin{tabular}{cccccc} 
 \toprule
 & Training parameter& \gemma &\gemmait& \mistral&\mistralit \\
 \midrule
\sparta\;/ \lora & epochs   & 14 / 6   & 1        &  1  &  1\\ 
              & batch size  & 40  & 40     &   40 &  40 \\ 
       		  &learning rate   &  1e-4 & 1e-4  &  5e-5 &  5e-5\\
% 	          &max grad norm &  -  & 60.0  &  80.0&  80.0\\
	          &dropout             &  0.0  & 0.05  &  0.1 &  0.1 \\
 	          &weight decay    &  0.0 & 0.0   &  0.0 &   0.0\\	    
 \bottomrule 
\end{tabular}
\caption{Training parameters used with both SpaRTA and LoRA for each of the pre-trained models in Table~\ref{tab:mmlu} (MMLU).} \label{tab:mmlu-training-params}
\end{table*}

\eat{
\begin{table*}[h]
\centering
% \footnotesize
\begin{adjustbox}{max width=\textwidth}
\begin{tabular}{lcccccccccccccc} 
\multicolumn{15}{c}{\textbf{\texttt{Model: \texttt{gemma-2b-it}}}} \\ 
\toprule
 & \multicolumn{2}{c}{\textbf{QNLI}} &  \multicolumn{2}{c}{\textbf{RTE}} &  
 \multicolumn{2}{c}{\textbf{SST2}}  &   \multicolumn{2}{c}{\textbf{QQP}} &
 \multicolumn{2}{c}{\textbf{MNLI}}  &   \multicolumn{2}{c}{\textbf{MRPC}} & \multicolumn{2}{c}{\textbf{COLA}} 
 \\ \cmidrule{2-15}
\multicolumn{1}{c}{}  & loss &acc.&  loss &acc.&  loss &acc.&  loss &acc.
    & loss &acc.&  loss &acc.&  loss &mcc\\ \midrule
\textbf{PT}  & 2.34 & 50.1 & 2.65 & 47.3 & 0.70 & 82.2 & 1.71 & 37.7 & 2.51 & 32.0 & 1.71 & 37.7 & 1.813 & -0.03 \\ \bottomrule \\
\multicolumn{15}{c}{\textbf{\texttt{Model: \texttt{Mistral-7B-Instruct-v0.3}}}} \\ 
\toprule
 & \multicolumn{2}{c}{\textbf{QNLI}} &  \multicolumn{2}{c}{\textbf{RTE}} &  
 \multicolumn{2}{c}{\textbf{SST2}}  &   \multicolumn{2}{c}{\textbf{QQP}} &
 \multicolumn{2}{c}{\textbf{MNLI}}  &   \multicolumn{2}{c}{\textbf{MRPC}} & \multicolumn{2}{c}{\textbf{COLA}} 
 \\ \cmidrule{2-15}
\multicolumn{1}{c}{}  & loss &acc.&  loss &acc.&  loss &acc.&  loss &acc.
    & loss &acc.&  loss &acc.&  loss &mcc\\ \midrule
\textbf{PT}  & 6.25 & 21.4 & 7.37 & 21.3 & 0.52 & 85.4 & 0.97 & 81.5 & 1.59 & 67.9 & 1.59  & 67.9 & 0.97 & 0.57 \\ \bottomrule
\end{tabular}
\end{adjustbox}
\caption{Zero-shot performance.} \label{tab:PT}
\end{table*}
}

\end{document}